\def\year{2021}\relax
\documentclass[letterpaper]{article} % DO NOT CHANGE THIS
\usepackage{aaai21}  % DO NOT CHANGE THIS
\usepackage{times}  % DO NOT CHANGE THIS
\usepackage{helvet} % DO NOT CHANGE THIS
\usepackage{courier}  % DO NOT CHANGE THIS
\usepackage[hyphens]{url}  % DO NOT CHANGE THIS
\usepackage{graphicx} % DO NOT CHANGE THIS
\usepackage{natbib}  % DO NOT CHANGE THIS AND DO NOT ADD ANY OPTIONS TO IT
\usepackage{caption} % DO NOT CHANGE THIS AND DO NOT ADD ANY OPTIONS TO IT
\usepackage{times}
\usepackage{latexsym}

\usepackage{multirow}
\usepackage[normalem]{ulem}
\useunder{\uline}{\ul}{}
\usepackage{numprint}
\usepackage{tabularx}
\usepackage{lipsum,adjustbox}
\usepackage{siunitx}
\usepackage{xcolor,colortbl}

\usepackage{tikz}
\tikzset{%
  every node/.append style={inner sep=5pt}
}
\usepackage{pgfplots}
\pgfplotsset{width=8cm,compat=1.9}
\usepgfplotslibrary{fillbetween}

\usepackage{microtype}

\title{Weight Squeezing: Reparameterization for Knowledge Transfer and Model Compression}

\author {
    Artem Chumachenko\textsuperscript{\rm 2 3}\thanks{\quad Equal contribution. List order determined in a Fortnite match.},
    Daniil Gavrilov\textsuperscript{\rm 1 2 3}\footnotemark[1],
    Nikita Balagansky  \textsuperscript{\rm 2 3},
    Pavel Kalaidin \textsuperscript{\rm 2 3 4}\thanks{\quad Work done while at VK.} \\
}
\affiliations {
    chumachenko.ad@phystech.edu, daniil.gavrilov@vk.com, balaganskij.nn@phystech.edu, p.kalaydin@tinkoff.ai\\
    \textsuperscript{\rm 1} VK, 
    \textsuperscript{\rm 2} VK Lab, 
    \textsuperscript{\rm 3} Moscow Institute of Physics and Technology, 
    \textsuperscript{\rm 4} Tinkoff

}

\begin{document}

\maketitle

\begin{abstract}

In this work, we present a novel approach for simultaneous knowledge transfer and model compression called \textbf{Weight Squeezing}. With this method, we perform knowledge transfer from a teacher model \textbf{by learning the mapping from its weights to smaller student model weights}.

We applied Weight Squeezing to a pre-trained text classification model based on BERT-Medium model and compared our method to various other knowledge transfer and model compression methods on GLUE multitask benchmark. We observed that our approach produces better results while being significantly faster than other methods for training student models.

We also proposed a variant of Weight Squeezing called Gated Weight Squeezing, for which we combined fine-tuning of BERT-Medium model and learning mapping from BERT-Base weights. We showed that fine-tuning with Gated Weight Squeezing outperforms plain fine-tuning of BERT-Medium model as well as other concurrent SoTA approaches while much being easier to implement.

% In this work, we present a novel approach for simultaneous knowledge transfer and model compression called \textbf{Weight Squeezing}. With this method, we perform knowledge transfer from a pre-trained teacher model \textbf{by learning the mapping from its weights to smaller student model weights}, without significant loss of model accuracy.

% We applied Weight Squeezing to a pre-trained text classification model and compared our method to various other knowledge transfer and model compression methods on several downstream text classification tasks based on the GLUE dataset. We observed that our approach produces better results than other methods for training student models without any loss in inference speed. We also compared Weight Squeezing with Low-Rank Factorization approach and observed that our method is significantly faster at inference while being competitive in terms of accuracy.

\end{abstract}
\section{Introduction}
\label{intro}

Today, deep learning has become a key technology in natural language processing (NLP), advancing state-of-the-art results for most NLP tasks. One of the significant achievements in applying deep learning in NLP is transfer learning. These methods include word vectors pre-trained on a large volume of text \cite{mikolov2013,glove2014}, which are now commonly used to initialize the first layer of neural networks for transfer learning. In recent times, methods such as ULMFiT \cite{ulmfit} have advanced the field of transfer learning in NLP by using language modeling during pre-training. Pre-trained language models now achieve state-of-the-art results on a diverse range of tasks in NLP, including text classification, question answering, natural language inference, coreference resolution, sequence labeling, and more \cite{qiu2020}.

Another breakthrough in NLP occurred after the introduction of Transformer \cite{transformer} neural networks that consist mostly of linear layers that compute the attention between model inputs. Unlike recurrent networks, Transformers have no recurrence over the spatial dimensions, allowing parallel computations, making it possible to significantly increase their size and reach state-of-the-art results for several tasks. \cite{devlin2019} introduced BERT (Bidirectional Encoder Representations from Transformers), a language representation model trained to predict masked tokens in texts from unlabeled data. Pre-trained BERT can be fine-tuned to create state-of-the-art models for a wide range of NLP tasks.

While BERT is capable of learning rich representations of text, using it for solving simple downstream tasks could be excessive. This is especially important when running downstream models on edge devices such as mobile phones. A common approach in such cases is model compression \cite{bertcomp}.

In this work, we present a novel approach to performing simultaneous transfer learning and model compression called \textbf{Weight Squeezing} where we learn the mapping from a teacher model's weights to a student model's weights.

We applied Weight Squeezing to the pre-trained teacher text classification model to obtain a smaller student model. We compared it with common model compression approaches, including variations of Knowledge Distillation without any reparameterizations and low-rank matrix factorization methods. Our experiments show that in most cases, Weight Squeezing achieves better performance than other baseline methods. 

We also proposed Gated Weight Squeezing to improve the accuracy of fine-tuning the BERT-Medium model, for which we combined fine-tuning with the mapping of larger BERT-Base weights. We showed that Gated Weight Squeezing produces higher accuracy than plain fine-tuning and compared trained model results with BERT, DistilBERT, TinyBERT, and MiniLM models.

\section{Motivation \& Problem Setting}
\label{motivation}

In this section we describe our motivation for building lightweight models.

Firstly, we focus on model size. For instance, a typical mobile app has the size of 100Mb, which significantly restricts the model size that can be used on the device\footnote{We managed to obtain models around 2.1 Mb in size, while the BERT-Medium model has a size of around 158 Mb.}. 
% restore after review
% For instance, VK\footnote{VK.com is an online social media platform with over 100 mln monthly users.} iOS app has the size of 100Mb, which significantly restricts us in the size of a model that we may use on the device\footnote{We manage to obtain models of size around 2.1 Mb while vanilla 12-layer BERT model is around 418 Mb in size.}. 
% For all our models, we have approximately 5Mb at maximum that we can use. 
Smaller models are also better for serving server-side.

Secondly, we paid particular attention to model inference time. A model may have fewer parameters but have a significant computational overhead regardless. Therefore, our focus is on making the student model faster than the teacher one.

Finally, we take into account that access to training resources, such as data and computing power, can be limited. Therefore, by focusing on task-specific compression, we limit the data we use to what we have for specific tasks only (see \ref{task} for details).

\section{Related Work}

\subsection{Transfer Learning \& Unsupervised Pre-training}

Self-supervised training as unsupervised pre-training became one of the key techniques for solving NLP tasks. While the amount of labeled data can be limited, we often have unlabeled data at our disposal. This data can be effectively utilized for pre-training some parts of the models. One way to conduct self-supervised pre-training of NLP models is language modeling. It can be performed using autoregressive models when learning to predict the next words in a text based on previous words.

% \subsection{Transformer}

% The Transformer was introduced as a type of neural network architecture for machine translation \cite{transformer}. Transformers consist of two parts: Encoder and Decoder. The former is used to read an original sentence to gather information for the latter, which outputs the translation as an autoregressive model conditioned on encoder outputs. Both Encoder and Decoder blocks have the same structure and consist of the same layers performing self-attention over the inputs. For a detailed description of the architecture, see \ref{appendix:architecture}.

\subsubsection{Transformer and BERT.}
The Transformer was introduced as a type of neural network architecture for machine translation \cite{transformer}. While the Transformer decoder can be used to train a language model, the performance of such a model could suffer because of its autoregressive nature. Masked Language Modeling (MLM) with BERT \cite{devlin2019} was proposed as an alternative to Language Modeling. This method involves training a model to predict masked words based on unmasked ones. The output of such a model depends on all words in the input text, which makes it capable of learning more complex data patterns.

\subsection{Model Compression}
The network architecture can be considerably over-parameterized and therefore inefficient in terms of memory and computing resources. Trading accuracy for performance and model size is often reasonable, which brings us to model compression techniques. There are many approaches \cite{bertcomp,qiu2020} to compressing BERT, including pruning \cite{poorman,fan2019, proximal,gordon,voita,qapruning}, quantization \cite{q8bert,qbert}, parameter sharing  \cite{albert}, and knowledge distillation. Some of these methods can be combined to achieve better results \cite{ladabert}.

% пойнт, что у нас не всегда есть доступ к датасету, на котором обучен BERT = например, quantization не подходит.

% % \subsubsection{Pruning}
% Pruning approaches remove less important parameters from the model.
% \cite{poorman,fan2019} explore several strategies to dropping layers. \cite{proximal,gordon,voita,qapruning} prune some parts of a model after training.

\subsubsection{Low-Rank Matrix Factorization.}
Low-rank matrix factorization approaches focus on reducing the size of model parameters. These approaches include Singular-Value Decomposition (SVD), Tensor Train (TT) Decomposition \cite{tt-svd}, and others. These methods can be considered a separate case of the pruning approach where we try to reduce the size of parameters by removing parts that can be considered unnecessary. For SVD, we drop some part of the weights which are related to small singular values, while TT can be seen as a generalization of SVD.

The most notable examples of using low-rank matrix factorization methods for NLP is reducing the embedding matrix size, which contains a significant part of model parameters \cite{albert, khrulkov2019, shu2019, acharya2019}.

\subsubsection{Knowledge Distillation.}
\label{kd}

In the Knowledge Distillation (KD) approach \cite{ba2014,hinton2015,romero2014fitnets} a smaller \textit{student} model is trained to mimic a \textit{teacher} BERT model. Current KD approaches for BERT can be categorized by what exactly they try to match as follows: distillation on encoder outputs/hidden states: \cite{extreme,tiny,mobile,pkd,distilbert}, distillation on model output \cite{extreme,pkd,distilbert,tiny,adabert}, distillation on attention maps \cite{mobile,tiny}.

KD can also be split into two categories: task-agnostic and task-specific.

\textbf{Task-agnostic KD}\label{task} involves reducing the size of BERT itself. Most methods fall under this type of compression. The KD methods, such as DistilBERT \cite{distilbert} or MobileBERT \cite{mobile} are trained on the same corpus as the one used when pre-training a BERT model from scratch. A typical scenario when using models such as BERT is to take a model already pre-trained on a very large corpus of data, since training or fine-tuning BERT is computationally expensive and could require a considerable amount of time. In some cases, even storing a large corpus, not to mention using it for training, could be a problem in this task.

% \subsubsection{Task-specific compression}
Instead of compressing BERT itself, a different approach of \textbf{task-specific KD} can be taken by fine-tuning BERT on a downstream task first and then applying compression techniques to train a smaller model. \cite{tp} use KD to train a single layer BiLSTM student model from BERT. \cite{dt} show that when given a large amount of unlabeled data, student BiLSTM networks can even match the performance of the teacher. \cite{turc2019} pre-train compact student models on unlabeled data and then apply KD.

\begin{center}
\begin{table*}
  \small
{\renewcommand{\arraystretch}{1.2}
\begin{tabularx}{\textwidth}{|X|c|r|cclccclccclccclc|}
\hline
% \multirow{5}{*}{} & \multicolumn{2}{l|}{} & \multicolumn{4}{l|}{NO-REPARAM.} & \multicolumn{4}{l|}{WS} & \multicolumn{4}{l|}{SVD} & \multicolumn{4}{l|}{TT} \\ \cline{2-19} 
 & \multicolumn{2}{r|}{Time CPU} & \multicolumn{2}{l|}{d16: $\times$1} & \multicolumn{2}{l|}{d32: $\times$1} & \multicolumn{2}{l|}{d16: $\times$1} & \multicolumn{2}{l|}{d32: $\times$1} & \multicolumn{2}{l|}{d16: $\times$5.3} & \multicolumn{2}{l|}{d32: $\times$4.9} & \multicolumn{2}{l|}{d16: $\times$89.2} & \multicolumn{2}{l|}{d32: $\times$126.8} \\ \cline{2-19} 
 & \multicolumn{2}{r|}{Time GPU} & \multicolumn{2}{l|}{d16: $\times$1} & \multicolumn{2}{l|}{d32: $\times$1} & \multicolumn{2}{l|}{d16: $\times$1} & \multicolumn{2}{l|}{d32: $\times$1} & \multicolumn{2}{l|}{d16: $\times$2.2} & \multicolumn{2}{l|}{d32: $\times$2.2} & \multicolumn{2}{l|}{d16: $\times$26.4} & \multicolumn{2}{l|}{d32: $\times$54.2} \\ \cline{2-19} 
 & \multicolumn{2}{r|}{Method} & \multicolumn{1}{c|}{MLE} & \multicolumn{2}{c|}{KD} & \multicolumn{1}{c|}{KD-EO} & \multicolumn{1}{c|}{MLE} & \multicolumn{2}{c|}{KD} & \multicolumn{1}{c|}{KD-EO} & \multicolumn{1}{c|}{MLE} & \multicolumn{2}{c|}{KD} & \multicolumn{1}{c|}{KD-EO} & \multicolumn{1}{c|}{MLE} & \multicolumn{2}{c|}{KD} & KD-EO \\ \cline{2-19} 
 & d & \multicolumn{1}{c|}{teacher} & \multicolumn{1}{l}{} & \multicolumn{1}{l}{} &  & \multicolumn{1}{l}{} & \multicolumn{1}{l}{} & \multicolumn{1}{l}{} &  & \multicolumn{1}{l}{} & \multicolumn{1}{l}{} & \multicolumn{1}{l}{} &  & \multicolumn{1}{l}{} & \multicolumn{1}{l}{} & \multicolumn{1}{l}{} &  & \multicolumn{1}{l|}{} \\ \hline \hline
% \multirow{2}{*}{SST2} & 32 & \multirow{2}{*}{91.4} & 82.7 & \multicolumn{2}{c}{82.1} & \multicolumn{1}{c|}{83.1} & 82.8 & \multicolumn{2}{c}{83.7} & \multicolumn{1}{c|}{83.8} & 0.0 & \multicolumn{2}{c}{0.0} & \multicolumn{1}{c|}{0.0} & 0.0 & \multicolumn{2}{c}{0.0} & 0.0 \\ \cline{2-2}
%  & 16 &  & 82.1 & \multicolumn{2}{c}{82.3} & \multicolumn{1}{c|}{82.9} & 84.1 & \multicolumn{2}{c}{82.9} & \multicolumn{1}{c|}{82.9} & 0.0 & \multicolumn{2}{c}{0.0} & \multicolumn{1}{c|}{0.0} & 0.0 & \multicolumn{2}{c}{0.0} & 0.0 \\ \hline
\multirow{2}{*}{MNLI} & 32 & \multirow{2}{*}{81.3} & 65.0 & \multicolumn{2}{c}{64.9} & \multicolumn{1}{c|}{70.5} & 71.5 & \multicolumn{2}{c}{64.4} & \multicolumn{1}{c|}{68.2} & 68.1 & \multicolumn{2}{c}{61.1} & \multicolumn{1}{c|}{70.9} & 68.2 & \multicolumn{2}{c}{67.3} & \textbf{72.6} \\ \cline{2-2}
 & 16 &  & 57.0 & \multicolumn{2}{c}{59.4} & \multicolumn{1}{c|}{59.2} & 57.3 & \multicolumn{2}{c}{57.3} & \multicolumn{1}{c|}{64.3} & 56.3 & \multicolumn{2}{c}{55.3} & \multicolumn{1}{c|}{60.8} & 62.1 & \multicolumn{2}{c}{62.0} & \textbf{65.4} \\ \hline \hline
\multirow{2}{*}{COLA} & 32 & \multirow{2}{*}{50.8} & 17.3 & \multicolumn{2}{c}{18.1} & \multicolumn{1}{c|}{17.4} & 17.0 & \multicolumn{2}{c}{19.3} & \multicolumn{1}{c|}{\textbf{20.7}} & 18.3 & \multicolumn{2}{c}{17.7} & \multicolumn{1}{c|}{18.0} & 16.0 & \multicolumn{2}{c}{17.4} & 14.7 \\ \cline{2-2}
 & 16 &  & 16.1 & \multicolumn{2}{c}{17.0} & \multicolumn{1}{c|}{15.6} & 15.0 & \multicolumn{2}{c}{16.6} & \multicolumn{1}{c|}{16.5} & 16.8 & \multicolumn{2}{c}{16.7} & \multicolumn{1}{c|}{16.0} & \textbf{18.2} & \multicolumn{2}{c}{17.7} & 17.5 \\ \hline \hline
% \multirow{2}{*}{STSB} & 32 & \multirow{2}{*}{87.0} & 21.0 & \multicolumn{2}{c}{20.9} & \multicolumn{1}{c|}{20.8} & 27.4 & \multicolumn{2}{c}{16.5} & \multicolumn{1}{c|}{28.1} & 27.6 & \multicolumn{2}{c}{28.8} & \multicolumn{1}{c|}{0.0} & 0.0 & \multicolumn{2}{c}{0.0} & 0.0 \\ \cline{2-2}
%  & 16 &  & 20.4 & \multicolumn{2}{c}{21.1} & \multicolumn{1}{c|}{0.0} & 17.9 & \multicolumn{2}{c}{0.0} & \multicolumn{1}{c|}{0.0} & 0.0 & \multicolumn{2}{c}{0.0} & \multicolumn{1}{c|}{0.0} & 0.0 & \multicolumn{2}{c}{0.0} & 0.0 \\ \hline
\multirow{2}{*}{MRPC} & 32 & \multirow{2}{*}{87.3} & 77.6 & \multicolumn{2}{c}{77.3} & \multicolumn{1}{c|}{77.9} & \textbf{79.0} & \multicolumn{2}{c}{78.5} & \multicolumn{1}{c|}{77.5} & 77.7 & \multicolumn{2}{c}{77.8} & \multicolumn{1}{c|}{78.2} & 77.7 & \multicolumn{2}{c}{77.8} & 78.5 \\ \cline{2-2}
 & 16 &  & \textbf{78.8} & \multicolumn{2}{c}{78.2} & \multicolumn{1}{c|}{78.0} & 78.1 & \multicolumn{2}{c}{78.5} & \multicolumn{1}{c|}{78.2} & 76.5 & \multicolumn{2}{c}{76.7} & \multicolumn{1}{c|}{77.5} & 78.7 & \multicolumn{2}{c}{78.5} & 78.5 \\ \hline \hline
% \multirow{2}{*}{QQP} & 32 & \multirow{2}{*}{88.4} & 85.2 & \multicolumn{2}{c}{0.0} & \multicolumn{1}{c|}{0.0} & 77.3 & \multicolumn{2}{c}{0.0} & \multicolumn{1}{c|}{0.0} & 0.0 & \multicolumn{2}{c}{0.0} & \multicolumn{1}{c|}{0.0} & 0.0 & \multicolumn{2}{c}{0.0} & 0.0 \\ \cline{2-2}
%  & 16 &  & 0.0 & \multicolumn{2}{c}{0.0} & \multicolumn{1}{c|}{0.0} & 0.0 & \multicolumn{2}{c}{0.0} & \multicolumn{1}{c|}{0.0} & 0.0 & \multicolumn{2}{c}{0.0} & \multicolumn{1}{c|}{0.0} & 0.0 & \multicolumn{2}{c}{0.0} & 0.0 \\ \hline
% \multirow{2}{*}{QNLI} & 32 & \multirow{2}{*}{89.2} & 61.3 & \multicolumn{2}{c}{61.6} & \multicolumn{1}{c|}{63.4} & 74.3 & \multicolumn{2}{c}{66.7} & \multicolumn{1}{c|}{79.9} & 0.0 & \multicolumn{2}{c}{0.0} & \multicolumn{1}{c|}{0.0} & 0.0 & \multicolumn{2}{c}{0.0} & 0.0 \\ \cline{2-2}
%  & 16 &  & 61.7 & \multicolumn{2}{c}{62.1} & \multicolumn{1}{c|}{61.9} & 64.6 & \multicolumn{2}{c}{64.6} & \multicolumn{1}{c|}{68.6} & 0.0 & \multicolumn{2}{c}{0.0} & \multicolumn{1}{c|}{0.0} & 0.0 & \multicolumn{2}{c}{0.0} & 0.0 \\ \hline
\multirow{2}{*}{RTE} & 32 & \multirow{2}{*}{70.0} & 59.2 & \multicolumn{2}{c}{59.6} & \multicolumn{1}{c|}{59.2} & 60.3 & \multicolumn{2}{c}{59.6} & \multicolumn{1}{c|}{59.6} & 60.7 & \multicolumn{2}{c}{\textbf{61.0}} & \multicolumn{1}{c|}{60.3} & 59.9 & \multicolumn{2}{c}{59.2} & 59.2 \\ \cline{2-2}
 & 16 &  & 58.5 & \multicolumn{2}{c}{58.8} & \multicolumn{1}{c|}{59.2} & 57.4 & \multicolumn{2}{c}{60.3} & \multicolumn{1}{c|}{\textbf{61.0}} & 58.5 & \multicolumn{2}{c}{58.5} & \multicolumn{1}{c|}{58.1} & 59.9 & \multicolumn{2}{c}{60.7} & 59.6 \\ \hline
\end{tabularx}
}
\caption{Accuracy on the GLUE tasks (see Section \ref{training_details} for further training procedure details). We also report inference time results (lower is better) for each of the reparameterization methods (see Section \ref{appendix:measurements}). We refer to $d$ as for the model hidden size (see Section \ref{appendix:low-rank} for the appropriate ranks for SVD and TT methods). See Table \ref{tab:speed} for the full list of speed measurements.}
  \label{acurracy-speed}
\end{table*}
\end{center}

\section{Weight Squeezing}

We now introduce a method to perform knowledge transfer and model compression by \textbf{learning the mapping between teacher and student weights}.

We start with a pre-trained teacher Transformer model with a large hidden state. It implies that for some linear layer $l$, we have a weight matrix $\Theta_l^t$ with the shape $n\times m$. 

We explore a case where the weights of a pre-trained teacher model are too big to run and store the model on an edge device. For this reason, we may want to train a student model with a smaller number of parameters. Let us say that we want the student model to make the weight matrix $\Theta_l^s$ at the same layer $l$ to have the shape equal to $a\times b$, where $a < n$ and $b < m$.

In this work, we propose reparameterizing student weights $\Theta_l^s$ as follows:

\begin{equation}
\label{eq:1}
\Theta^{s} = \mathcal{L}\Theta^t\mathcal{R}
\end{equation}

where $\mathcal{L}$ and $\mathcal{R}$ are randomly initialized trainable parameters of the mapping with shapes equal to $a\times n$ and $m\times b$ respectively (here and below, we omitted the $l$ subscript for simplicity).

In this approach, instead of training student model weights from scratch, we reparameterize them as a trainable linear mapping from teacher model weights. Doing so allows us to transfer knowledge stored in the teacher weights to the student weights.

At the same time, mapping of teacher biases and word embeddings is performed as a single linear mapping as follows:

\begin{equation}
\label{eq:2}
\Theta^{s}_{single} = \Theta^t\mathcal{R}
\end{equation}

where biases are matrices of size $1 \times b$ and word embeddings have size $V \times b$, and $V$ is the total number of words in the vocabulary. This reparameterization for word embeddings can be seen as a linear alignment from pre-trained embeddings.

After reparameterization of the student model weights using Equations \ref{eq:1} and \ref{eq:2}, we train mapping weights $\mathcal{L}$ and $\mathcal{R}$ using plain negative log-likelihood (Weight Squeezing) or KD loss (Weight Squeezing combined with KD). When the mapping weights are trained, we compute student weights and then use them to make predictions dropping $\mathcal{L}$ and $\mathcal{R}$ matrices.

\subsection{Applications}

The proposed method defines the general way to use weights of one model to train another. In this paper, we concentrate on two ways of applying Weight Squeezing.

\subsubsection{Extreme Model Compression}
In many cases, we may want to obtain substantially smaller models (see Section \ref{motivation}). However, the list of available pre-trained models that can be used for fine-tuning is limited. Thus, if the target model size is very small, it is necessary to either train a new smaller BERT model or perform a task-specific compression of a larger model (see Section \ref{kd}).

In this work, we concentrate on applying Weight Squeezing for task-specific model compression. For this purpose, we fine-tuned the BERT-Medium model (41M parameters) on a particular dataset to obtain the pre-trained teacher model. We then applied Weight Squeezing to reparameterize weights of the significantly smaller target model (1M and 0.5M parameters).

\subsubsection{Fine-Tuning a pre-trained model}
While Weight Squeezing can be used for extreme model compression when no pre-trained smaller BERT is available, it is also suitable for fine-tuning BERT.

In this setup, we may want to fine-tune a BERT model (in our experiments we used BERT-Medium) on a specific task. A typical way to do so is to initialize a new model with pre-trained weights and then train it on a particular dataset. However, larger BERTs (e.g., BERT-Base) are available, and knowledge from these models could also be utilized during the fine-tuning.

In this work we propose \textbf{Gated Weight Squeezing}. In this case we fine-tuned BERT-Base on specific task to obtain large teacher network and then reparameterize weights of student model as follows:

\begin{equation}
\label{eq:3}
\Theta^{s} = \left(1 - \sigma(s)\right) \odot \mathcal{L}\Theta^t\mathcal{R} + \sigma(s) \odot \Theta^b
\end{equation}

$\Theta^t$ are weights of the teacher model (which are fine-tuned BERT-Base model), $\Theta^b$ are weights of BERT-Medium, $s$ is a scalar value, and $\sigma$ is the sigmoid function. We used $\mathcal{L}$, $\mathcal{R}$, $\Theta^b$, and $s$ as a trainable parameters. Embeddings are also reparameterized in a gated way following Equation \ref{eq:2}.
\subsection{Comparison to Similar Methods}
\subsubsection{Knowledge Distillation.}

Weight Squeezing (WS) is orthogonal to Knowledge Distillation methods. One could reparametrize weights of student model with WS and then train it with the arbitrary training method, such as Knowledge Distillation or plain Likelihood Maximization (MLE).

\subsubsection{DistilBERT and other compressed BERTs}
DistilBERT is orthogonal to models compressed with WS, just like KD is orthogonal to WS. Note that the typical way to compress BERT is task-agnostic compression (see Section \ref{kd}) for which we firstly obtain a smaller general BERT model trained on MLM task, and then use this smaller model for specific tasks. While in this paper, we concentrate on task-specific compression when we compress a large BERT model concerning the task. However, one could apply Weight Squeezing in pair with Knowledge Distillation to compress BERT itself. Nevertheless, we compare the proposed approach with variants of compressed BERT.

\subsubsection{Low-Rank Matrix Factorization.}
\label{low-rank-critique}
% It is important to note that Low-rank matrix factorization methods lose information about teacher weight at the beginning of the training procedure. I.e., if we take some weight, apply SVD to it and take $n$ ranks with the biggest singular values, then we drop other ranks that are no longer used in the training procedure. For WS, we use full teacher weight during the training to find its most important parts.

We observed that factorization methods could be hard to train and evaluate due to their large memory and computation footprint. Regardless of the factorization rate $r$, the result after applying the factorized layer will always retain its original size. Thus, some operations in the Transformer are not going to benefit from the model weights' factorization. For example, suppose we have a Transformer with the hidden layer's size equal to $1024$. Then even with small rank value $r$, self-attention will be computed between vectors with shape $1024$ \footnote{While the standard Transformer has the time complexity of attention layer equal to $\mathcal{O}(n^{2}h)$, where $n$ is sequence length and $h$ is the size of the hidden vector. For more modern Transformers \cite{lin-tran}, this complexity could be replaced with $\mathcal{O}(nh^2)$ which is significantly more preferable for $h$ smaller than $n$ (which is often to occur). Thus models that do not rely on Low-Rank factorization could further benefit from reducing the hidden size, making them even faster than SVD or TT factorized models.}. Because of this, very low rates of $r$ did not lead to a performance boost.

It is also important to note that there is often a practical limit of Low-Rank methods' compression rate (e.g., factorization rank equal to $1$ for SVD), which often easy to reach if the teacher model is big enough.

\section{Model Compression with Weight Squeezing}
\label{experiments}

\subsection{Baselines}
\label{baselines}
We trained all models on GLUE datasets \cite{glue}.

For each dataset, we trained a teacher model by fine-tuning the pre-trained BERT-Medium model. 

In this paper, we consider the following methods for reparameterization of student models:
\begin{enumerate}
    \item No weight reparametrization
    \item Weight Squeezing (WS) of the teacher model
    \item SVD applied to the teacher model
    \item TT applied to the teacher model
\end{enumerate}

Each of the baselines above could be trained with ambiguous methods. We used the following approaches:
\begin{enumerate}
    \item Maximum Likelihood Estimation (MLE)
    \item Knowledge Distillation (KD) of the teacher model
    \item Knowledge Distillation on Encoder Outputs (KD-EO) of the teacher model
\end{enumerate}

Thus, for some of the baselines (e.g., WS with KD), we used the teacher model for weight reparameterization and getting teacher predictions to estimate the loss.

Since we focused on making models smaller in terms of the overall number of parameters, we trained student models in two configurations of small hidden sizes equal to $16$ and $32$. For all models, we used the number of heads equal to $4$ and used the same with the teacher number of Transformer layers equal to $8$.

\begin{table}[h]
\small
{\renewcommand{\arraystretch}{1.4}
\centering
\begin{tabularx}{235pt}{|X|c|c|c|c|c|}
\hline
\multicolumn{1}{|c|}{SA Heads:}    & \multicolumn{2}{c|}{4} & 8     \\ \hline
\multicolumn{1}{|c|}{Hidden size:} & 16         & 32        & 512   \\ \hline
Plain BERT                         & 0.52M      & 1.1M      & 41.3M \\ \hline
WS                                 & 4.18M      & 8.4M      & -     \\ \hline
SVD                                & 0.53M      & 1.1M      & -     \\ \hline
TT                                 & 0.53M      & 1.1M      & -     \\ \hline

\end{tabularx}
}
\caption{The number of parameters for each model. Note that once the WS model is trained, we no longer have to store mappings weights; thus, WS will have the number of parameters equal to Plain BERT during the inference.}
\label{tab:parameters}
\end{table}

\subsection{Reparameterization methods}
\subsubsection{No weight reparametrization} 
For this experiment, we randomly initialize parameters of a student model without utilizing teacher model in weight reparameterization.

\subsubsection{Weight Squeezing} 
For Weight Squeezing, we used fine-tuned teacher models as the source of mapping for weights reparameterization. This way, we reparameterized the parameters of all linear layers in the model as in Equation \ref{eq:1} and embedding vectors as in Equation \ref{eq:2}. Weights of the mapping for linear layers were initialized as Xavier Normal, while the mapping for the embedding matrix was initialized as Xavier Uniform. We optimized the loss with respect to the mapping parameters used to reparameterize the student model weights and the rest of the student model parameters that were not reparameterized (e.g., the layer normalization weights).

\subsubsection{Low-Rank Matrix Factorization.}
\label{appendix:low-rank}

We also experimented with the matrix factorization approaches (LRMF) consisted of two methods: \textbf{SVD} and \textbf{TT}

For \textbf{SVD} we started with factorizing teacher model weights as follows
\[\Theta = U_{m\times m}\Sigma_{m\times n}V^{\top}_{n\times n}\]

where $\Sigma$ is a diagonal matrix of the singular values, and $U$ and $V$ are the left and right singular vectors of the weight.

By keeping only $r$ largest singular values, we could obtain the reduced form of this weight :

\[\hat{\Theta} = U_{m\times r}\Sigma_{r\times r}V^{\top}_{r\times n} \approx \mathcal{U}_{m\times r}V^{\top}_{r\times n}\]

Thus instead of storing $nm$ parameters for each weight, we would have to keep only $r(n + m)$ parameters. If $r$ value is small enough, then the total number of parameters will be reduced compared to the original model.

In our experiments, we applied SVD factorization of the teacher weights and then trained $\mathcal{U}_{m\times r}$ and $V_{r\times n}$ to minimize the loss.

We also applied \textbf{TT} to obtain the reduced form of the teacher weights with $4$ cores and also trained this cores to minimize the loss.

Note that in Low-Rank approaches, we do not directly train the student model with the specified hidden state size as in Non Low-Rank Factorization methods (Non-LRFM). Instead, we injected a bottleneck in the middle of each layer, which allowed us to reduce the total number of parameters in the model. For this reason we evaluated the number of parameters for Non-LRFM models for hidden sizes equal to $16$ and $32$ and then found appropriate factorization ranks to make SVD and TT models have approximately similar number of parameters. Thus, we compared Non-LRFM models of hidden size equal to $16$ and $32$ with SVD model with $r$ equal to $2$ and $7$, and TT model with rank equal to $9$ and $18$ respectively (See Table \ref{tab:parameters}).

\begin{table}[h]
\small
{\renewcommand{\arraystretch}{1.4}
\centering
\begin{tabularx}{0.5\textwidth}{|X|p{2cm}|p{2cm}|p{2.4cm}|}
\hline
\multicolumn{4}{|c|}{CPU}  \\ \hline \hline
Size & Non-LRMF & SVD & TT \\ \hline
512 & 8603 $\pm$ 663 ms & - & - \\ \hline
32 & \textbf{531 $\pm$ 14 ms} \newline ($\times$1) & 2598  $\pm$ 62 ms \newline ($\times$4.9) & 67332 $\pm$ 11146 ms \newline ($\times$89.2) \\ \hline
16 & \textbf{466 $\pm$ 16 ms} \newline ($\times$1) & 2488 $\pm$ 114 ms \newline ($\times$5.3) & 41554 $\pm$ 325 ms \newline ($\times$126.2) \\ \hline \hline

\multicolumn{4}{|c|}{GPU}  \\ \hline \hline
Size & Non-LRMF & SVD & TT \\ \hline
512 & 2667 $\pm$ 49 ms & - & - \\ \hline
32 & \textbf{505 $\pm$ 38 ms} \newline ($\times$1) & 1122 $\pm$ 2 ms \newline ($\times$2.2) & 27382 $\pm$ 22 ms \newline ($\times$26.4) \\ \hline
16 & \textbf{517 $\pm$ 49 ms} \newline ($\times$1) & 1115 $\pm$ 2 ms \newline ($\times$2.2) & 13659 $\pm$ 12 ms \newline ($\times$54.2) \\ \hline

\end{tabularx}
}
\caption{Inference speed results for Non-Low Rank Matrix Factorization and decomposed with SVD and TT  models. \textbf{Non-LRFM include No-reparameterization and WS methods}. We compare models of comparable size. '-' means no experiments were conducted.}
\label{tab:speed}
\end{table}

\subsection{Training objectives}
\subsubsection{Maximum Likelihood Estimation}
In this method we utilize plain MLE for model training
\begin{equation}
\label{eq:4}
\mathcal{L}_{MLE} = -\log(p^s_c)
\end{equation}

where $p^s$ is the student model's output probabilities, $p^s_i$ is the $i$-th component of student predictions, and $c$ is the ground truth label index. 

For this method, we do not use predictions of the teacher model to train on. Thus, a training model with MLE and no weight reparametrization setting could be seen as a training student model from scratch without any knowledge transfer methods.

\subsubsection{Knowledge Distillation}

For KD we also utilize predictions of the teacher model in the training:

\begin{equation}
\label{eq:5}
\mathcal{L}_{KD} = -\alpha \log(p^s_c) - (1 - \alpha) \sum_i p^t_i \log(p^s_i)
\end{equation}

The notation follows Equation \ref{eq:4} adding $i$-th component of the teacher model prediction $p^t_i$. Thus, the first term in the Equation \ref{eq:5} is a negative log-likelihood used for MLE, while the second stands for part of KL-Divergence between teacher and student we could optimize with respect to student parameters. 

We also used temperature to evaluate KL-Divergence by dividing prediction logits of both teacher and student models by the same number before applying softmax.

\subsubsection{Knowledge Distillation on Encoder Outputs}
\label{kd-eo}
With this loss, we also utilize hidden states of the teacher model during the training procedure.
The resulting loss function is defined as:

\[
\mathcal{L}_{KD-EO} = -\alpha \log p^s_c - \beta \sum_i p^t_i \log p^s_i + \gamma \sum_j L2(h^t_j, f_j(h^s_j))
\]

where the first two terms in the equation above correspond to terms in Equation \ref{eq:5}, while the last term is the mean squared error between teacher hidden state $h^t_j$ from layer $j$ and linearly mapped student hidden state $f_j(h^s_j)$ from the same layer. We defined $f_j$ as a fully connected layer and optimized this loss with respect to the student model parameters and each mapping $f_j$. 

Note that $\alpha + \beta + \gamma = 1$. Also, since there are several hidden states in each layer corresponding to different words in the input sequence, we used the first hidden state to evaluate the loss function.

\subsection{Training details}
\label{training_details}
Hyperparameters for each model were found using Bayesian hyperparameter search (see  Table \ref{tab:hyp} for ranges of the search). We performed a search for about $8$-$15$ GPU days for each model with NVidia A100 GPU. We maximized the appropriate metric on each GLUE dataset dev split to find each method's best training configuration and report the best performing method's results.

We used Adam \cite{adam} optimizer to train all models with a linear warmup and linear decay of a learning rate. We also applied dropout to the attention matrix and to averaged hidden state before the last linear layer, which produced logits of predictions.

% Since models on STSB dataset are trained on regression task, we replaced  KL-Divergence in $\mathcal{L}_{KD}$ and $\mathcal{L}_{KD-EO}$ losses with L2 distance between teacher and student predictions. In this case, the temperature was not used to evaluate the loss.

\begin{table}[h]
\small
{\renewcommand{\arraystretch}{1.4}
\centering
\begin{tabularx}{235pt}{|X|l|}

\hline
\multicolumn{2}{|c|}{General} \\ \hline
learning rate                        & $[1$e-$3, 5$e-$4, 3$e-$4, 1$e-$4, 5$e-$5, 3$e-$5, 3$e-$5]$ \\ \hline
beta1                                & $[ 0.7, 0.8, 0.9, 0.99, 0.999 ]$                             \\ \hline
beta2                                & $[ 0.9, 0.99, 0.999 ]$                                       \\ \hline
warmup steps                         & $[ 100, 500, 1000, 2000, 4000, 8000 ]$                       \\ \hline
batch size                           & $[ 8, 16, 32, 64, 128 ]$                                     \\ \hline
hidden dropout                       & $[ 0.1, 0.15, 0.2, 0.25, 0.3 ]$                              \\ \hline
attention dropout                    & $[ 0.1, 0.15, 0.2, 0.25, 0.3 ]$                              \\ \hline \hline
\multicolumn{2}{|c|}{Knowledge Distillation}                                                        \\ \hline
$\alpha$                             & $[0;1]$                                                      \\ \hline
temperature                       & $[1, 2, 3, 4, 5, 6, 7, 8, 9, 10]$                            \\ \hline \hline
\multicolumn{2}{|c|}{Knowledge Distillation Encoder Outputs}                                        \\ \hline
$f_j$ learning rate                        & $[1$e-$3, 5$e-$4, 3$e-$4, 1$e-$4, 5$e-$5, 3$e-$5, 3$e-$5]$ \\ \hline
$\alpha$ logit                       & $[-5; 5]$                                                    \\ \hline
$\beta$ logit                        & $[-5; 5]$                                                    \\ \hline
$\gamma$ logit                       & $[-5; 5]$                                                    \\ \hline
temperature                        & $[1, 2, 3, 4, 5, 6, 7, 8, 9, 10]$                            \\ \hline \hline
\multicolumn{2}{|c|}{Gated Weight Squeezing}                                        \\ \hline
Initial gate $s$                        & $[1; 4]$ \\ \hline
\end{tabularx}
}
\caption{Hyperparameter search ranges for methods used for hyperparameter search.
}

  \label{tab:hyp}
\end{table}

\subsection{Inference Speed Measurements}
\label{appendix:measurements}

We compared Non-Low Rank Matrix Factorization models (i.e., no weight reparameterization and weight squeezing) with models factorized using the TT and SVD methods for inference speed measurement.

In our experiments, we used the same models as in other experiments (see Sections \ref{baselines} and \ref{appendix:low-rank} for Non-LRMF parameters and for parameters of TT and SVD approaches). All models have a comparable number of parameters (see Table \ref{tab:parameters}).

We evaluated models on sequences with input lengths of $128$ (see Table \ref{tab:speed}). We ran the models on $1000$ samples with batch size equal to $16$ for GPU measurements and $100$ samples with batch size equal to $1$  for CPU. We repeated the measurements $5$ times and reported the mean and std values of the models' total computation time. For CPU measurements, we used 1.8 GHz Intel DualCore i5, while for GPU measurements, we used NVIDIA Tesla T4.

All models were prepared with PyTorch JIT compilation since we found that it slightly increases all models' speed in this experiment.

\begin{center}
\begin{table*}
  \small
  \centering
{\renewcommand{\arraystretch}{1.2}}
\begin{tabularx}{500pt}{|X|l|l|l|l|l|l|l|l|l|l|l|}
\hline
\textbf{Model} & \textbf{\#Param} & \textbf{SST2} & \textbf{CoLA} & \textbf{STSB} & \textbf{MRPC} & \textbf{QQP} & \textbf{QNLI} & \textbf{RTE} & \textbf{MNLI} & \textbf{RACE} & \textbf{SQuAD2} \\ \hline \hline
BERT-base       & 109M             & 93.2          & 58.9          & -             & 87.3          & 91.3         & 91.7          & 68.6      &  84.5 & - & 76.8 \\ \hline
BERT-medium     & 41M              & 91.4          & 48.6          & 87.0          & 87.3          & 88.4         & 89.2          & 70.0        &  & & \\ \hline \hline
DistilBERT    & 66M              & 90.7          & 43.6          & -             & 87.5          & 84.9         & 85.3          & 59.9      &  79.0 & &  70.7 \\ \hline \hline
TinyBERT       & 66M              & 91.6          & 42.8          & -             & 88.4          & 90.6         & 90.5          & 72.2       &  83.5 & - & 73.1 \\ \hline
TinyBERT       & 17M              & 88.4          & -             & -             & -             & -            & -             & -          & 77.4 & - & 63.6 \\ \hline
\hline
MiniLM         & 66M              & 92.0          & 49.2          & -             & 88.4          & 91.0         & 91.0          & 71.5         & 84.0 & - & 76.4\\ \hline
MiniLM         & 17M              & 89.7          & -             & -             & -             & -            & -             & -            & 79.1 & - & 66.9\\ \hline \hline

ALBERTbase-128         & 12M              & 90.3          & -          & -             & -         & -         & -          & -          &  81.6 & 64.0 & 80.0/77.1 \\ \hline
ALBERTbase-64         & 10M              & 89.4          & -          & -             & -         & -         & -          & -          &  80.8 & 63.5 & 77.5/74.8 \\ \hline \hline

WS-large          & 10M             &           &               &           &           &              &               &          &  & &\\ \hline
WS-32          & 1.1M             & 83.8          &               & 28.1          & 79.0          &              &               & 60.3         & 71.5 & &\\ \hline
WS-16          & 0.52M            & 84.1          &               & 17.9          & 78.5          &              &               & 61.0         &  64.3 & &\\ \hline
\end{tabularx}
\caption{Accuracy accross benchmark tasks.}
\end{table*}
\end{center}

\section{Fine-Tuning with Gated Weight Squeezing}
\label{gated-weight-squeezing}

A common practice for training a model on a specific task using pre-trained BERT is to initialize the model with this pre-trained state and then fine-tune it.

In this experiment, we compared Gated Weight Squeezing (41M parameters) with widely adopted models for fine-tuning on different GLUE tasks:
\begin{enumerate}
    \item BERT-Medium (41M)
    \item DistilBERT (66M)
    \item TinyBERT (66M)
    \item MiniLM (66M)
\end{enumerate}

\subsection{Training Details}

While BERT-Medium has 41M parameters, larger models are available (e.g., BERT-Base with 109M parameters). These larger models could be used during the fine-tuning procedure to training more accurate models for a specific task.

To do so, we firstly trained a teacher model by fine-tuning BERT-Base. Then we used this fine-tuned model and BERT-Medium for the reparameterization of student model weights proposed in the Equation \ref{eq:3}, which results in the student model with 41M parameters, which we compared with baseline fine-tuning. The reparameterized student model was trained with KD-EO loss using the BERT-Base as a teacher (see Section \ref{kd-eo}).

Since BERT-Base and BERT-Medium have a different number of hidden layers ($12$ and $8$ respectively), we used the first $8$ layers of BERT-Base to perform mapping of the weights. We also used hidden states of the first $8$ layers to evaluate the KD-EO loss.

We followed the same training strategy as in Section \ref{training_details}. We added a new hyperparameter for training the Gated WS model for the initial gate $s$ value, which we used to select from the range equal to $[1; 4]$.

Note that the usual fine-tuning of BERT-Medium could be seen as a special case of Gated Weight Squeezing with a fixed $\sigma(s) = 1$.

\section{Results}

See Table \ref{acurracy-speed} for the list of best performing model accuracies. Inference speed measurements of baselines can be found in Table  \ref{tab:speed}.

We observed that WS generally produced better results than models trained without weight reparameterization for knowledge transfer. SVD and TT methods were competitive to WS since LRFM operates with substantially bigger hidden states to evaluate the attention. However, SVD and TT models were significantly slower in inference than Non-LRFM (\textbf{2-5} times slower for SVD and \textbf{26-127} times for TT), which makes them be an inappropriate choice in cases when speed does matter. Also, note that in most cases, when the best result for some dataset was produced by SVD or TT model, the second-best performing result was achieved by Weight Squeezing, while the latter is significantly faster (e.g., as for MNLI dataset).

Since we performed hyperparameters search for all models for the same amount of time, models trained with Knowledge Distillation and Knowledge Distillation on Encoder Outputs losses produced lower accuracy compared to MLE loss in some cases, since they imply searching appropriate $\alpha$, $\beta$, and $\gamma$ values for training. We observed that the training algorithm is sensitive to these parameters, making KD loss difficult to use if the training process takes a long time, and it is hard to perform a comprehensive hyperparameter search procedure.

\subsection{Model Size}
We compared the number of baseline model parameters (See Table \ref{tab:parameters}). The Plain BERT row stands for the number of parameters in the ordinary classifier built on top of the BERT model with a specific number of layers and self-attention heads.

For the TT and SVD methods, we show the number of parameters for the specific factorization rates used to make these approaches have the number of parameters approximately equal to the plain classifier (see \ref{appendix:low-rank} for more details). 

Note that \textbf{after the WS model is trained, we no longer have to evaluate the mapping result}. The inference setting parameters are equal to the Plain BERT row for the appropriate hidden sizes.

\section{Conclusion \& Future Work}

We introduced Weight Squeezing, a novel approach to knowledge transfer and model compression. We showed that it could compress pre-trained text classification models and create compelling lightweight and fast models. 

We showed that Weight Squeezing successfully utilized teacher weights and produced better results than models trained without weight reparameterization. We also showed that Weight Squeezing could be a competitive alternative to Low-Rank Factorizing Methods in terms of accuracy, but being significantly faster.

While the current work focused on transferring knowledge to task-specific models, we would like to apply Weight Squeezing for task-agnostic compression to create more applicable BERT models trained for Masked Language Modelling tasks.

We are currently experimenting with the initialization of mappings and plan to continue this research. We are interested in applying this method in domains beyond NLP to compress other types of layers (convolutional, etc.). Also, one may find it important to reduce memory footprint during Weight Squeezing training, making such mappings of teacher weights to make training procedure more efficient.
% We introduced Weight Squeezing, a novel approach to knowledge transfer and model compression. We showed that it can be used to compress pre-trained text classification models and create compelling lightweight and fast models. We showed this approach could be a competitive alternative to Knowledge Distillation methods.

% While the current work was focused on transferring knowledge to task-specific models, we would like to apply Weight Squeezing for task-agnostic compression for creating more applicable BERT models trained for tasks such as Masked Language Modelling.

% We are currently experimenting with the initialization of mappings and plan to continue this research.

% We are also interested in applying this method in domains beyond NLP to compress other types of layers (convolutional, etc.).

% \section{Acknowledgements}


\bibliography{aaai21}
\clearpage

\end{document}